\documentclass[conference]{IEEEtran}
\IEEEoverridecommandlockouts
\usepackage{cite}
\usepackage{amsmath,amssymb,amsfonts}
\usepackage{algorithmic}
\usepackage{multirow}
\usepackage{graphicx}
\usepackage{hyperref}
\usepackage{textcomp}
\usepackage[super]{nth}
\usepackage{xcolor}
\def\BibTeX{{\rm B\kern-.05em{\sc i\kern-.025em b}\kern-.08em
    T\kern-.1667em\lower.7ex\hbox{E}\kern-.125emX}}
\begin{document}

\title{DeepFakes Evolution: Analysis of Facial Regions \\ and Fake Detection Performance
}


\author{Ruben Tolosana, Sergio Romero-Tapiador, Julian Fierrez and Ruben Vera-Rodriguez \\
Biometrics and Data Pattern Analytics - BiDA Lab, Universidad Autonoma de Madrid \\
{\tt\small \{ruben.tolosana, julian.fierrez, ruben.vera\}@uam.es, sergio.romerotapiador@gmail.com}
}

\maketitle

\begin{abstract}
Media forensics has attracted a lot of attention in the last years in part due to the increasing concerns around DeepFakes. Since the initial DeepFake databases from the \nth{1} generation such as UADFV and FaceForensics++ up to the latest databases of the \nth{2} generation such as Celeb-DF and DFDC, many visual improvements have been carried out, making fake videos almost indistinguishable to the human eye. This study provides an exhaustive analysis of both \nth{1} and \nth{2} DeepFake generations in terms of facial regions and fake detection performance. Two different methods are considered in our experimental framework: \textit{i)} the traditional one followed in the literature and based on selecting the entire face as input to the fake detection system, and \textit{ii)} a novel approach based on the selection of specific facial regions as input to the fake detection system.

Among all the findings resulting from our experiments, we highlight the poor fake detection results achieved even by the strongest state-of-the-art fake detectors in the latest DeepFake databases of the \nth{2} generation, with Equal Error Rate results ranging from 15\% to 30\%. These results remark the necessity of further research to develop more sophisticated fake detectors.
\end{abstract}


\section{Introduction}

Fake images and videos including facial information generated by digital manipulations, in particular with DeepFake methods~\cite{tolosana2020SurveyFakes,verdoliva2020media}, have become a great public concern recently~\cite{TED_news_concerns,BBC_doubleVideos}. The very popular term ``DeepFake" is referred to a deep learning based technique able to create fake videos by swapping the face of a person by the face of another person. Open software and mobile applications such as ZAO\footnote{\url{https://apps.apple.com/cn/app/id1465199127}} allow nowadays to automatically generate fake videos by anyone, without a prior knowledge of the task. But, how real are these fake videos compared with the authentic ones\footnote{\url{https://www.youtube.com/watch?v=UlvoEW7l5rs}}? 

Digital manipulations based on face swapping are known in the literature as Identity Swap, and they are usually based on computer graphics and deep learning techniques~\cite{tolosana2020SurveyFakes}. Since the initial publicly available fake databases, such as the UADFV database~\cite{li2018ictu}, up to the recent Celeb-DF and Deepfake Detection Challenge (DFDC) databases~\cite{li2019celebdf,dolhansky2019deepfake}, many visual improvements have been carried out, increasing the realism of fake videos. As a result, Identity Swap databases can be divided into two different generations.

In general, fake videos of the \nth{1} generation are characterised by: \textit{i)} low-quality synthesised faces, \textit{ii)} different colour contrast among the synthesised fake mask and the skin of the original face, \textit{iii)} visible boundaries of the fake mask, \textit{iv)} visible facial elements from the original video, \textit{v)} low pose variations, and \textit{vi)} strange artifacts among sequential frames. Also, they usually consider controlled scenarios in terms of camera position and light conditions. Many of these aspects have been successfully improved in databases of the \nth{2} generation. For example, the recent DFDC database considers different acquisition scenarios (i.e., indoors and outdoors), light conditions (i.e., day, night, etc.), distances from the person to the camera, and pose variations, among others. So, the question is, how easy is for a machine to automatically detect these kind of fakes?

\begin{figure*}[t]
\begin{center}
   \includegraphics[width=0.98\linewidth]{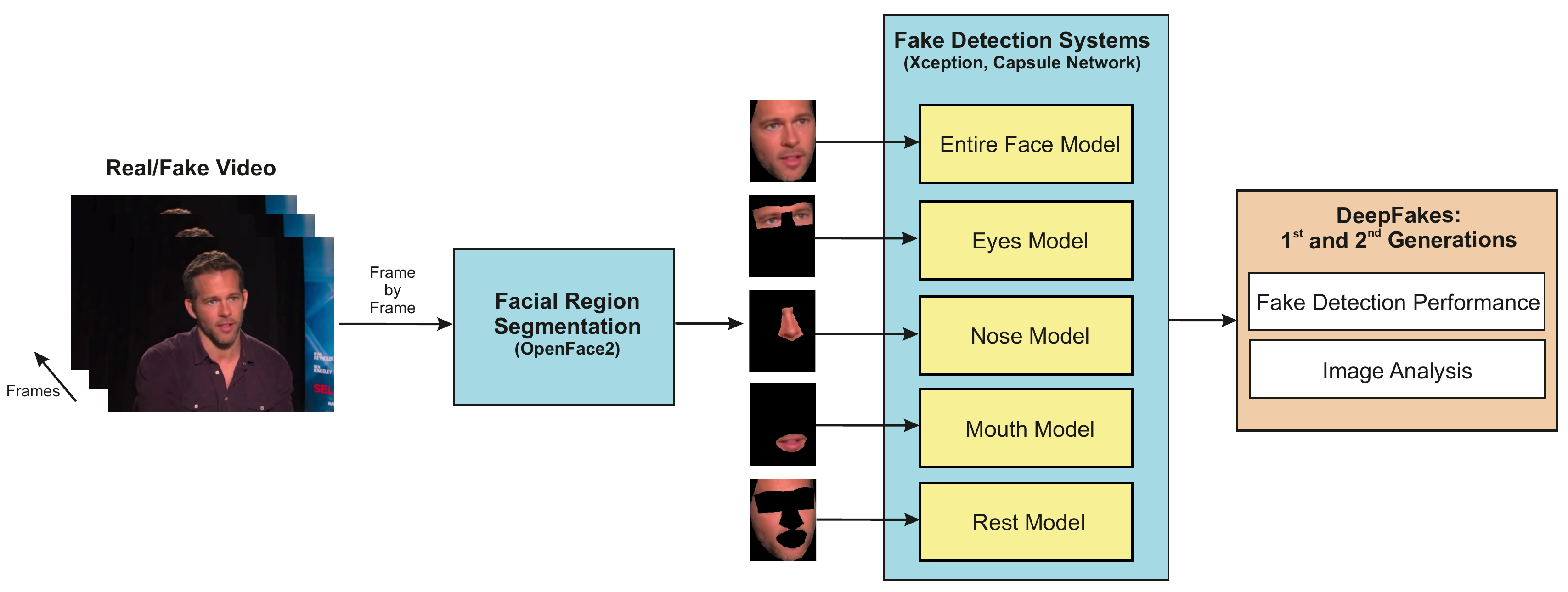}
\end{center}
   \caption{Architecture of our evaluation framework to analyse both facial regions and fake detection performance in DeepFake video databases of the \nth{1} and \nth{2} generations. Two different approaches are studied: \textit{i)} selecting the entire face as input to the fake detection system, and \textit{ii)} selecting specific facial regions.}
\label{fig:abstract_figure}
\end{figure*}

Different fake detectors have been proposed based on the visual features existed in the \nth{1} generation of fake videos. Yang \textit{et al.} performed in~\cite{yang2019exposing} a study based on the differences existed between head poses using a full set of facial landmarks (68 extracted from DLib~\cite{king2009dlib}) and those in the central face regions to differentiate fake from real videos. Once these features were extracted, Support Vector Machines (SVM) were considered for the final classification, achieving an Area Under the Curve (AUC) of 89.0\% for the UADFV database~\cite{li2018ictu}. 

The same authors proposed in~\cite{Li2019CVPR} another approach based on the detection of face warping artifacts. They proposed a detection system based on Convolutional Neural Networks (CNN) in order to detect the presence of such artifacts from the face and the surrounding areas. Their proposed detection approach was tested using the UADFV and DeepfakeTIMIT databases~\cite{li2018ictu,korshunov2018deepfakes}, outperforming the state of the art with 97.4\% and 99.9\% AUCs, respectively. 

Agarwal \textit{et al.} proposed in~\cite{2019_Agarwal} a detection technique based on facial expressions and head movements. Their proposed approach achieved a final AUC of 96.3\% over their own database, being robust against new manipulation techniques. 

Finally, Sabir \textit{et al.} proposed in~\cite{2019_Sabir} to detect fake videos through the temporal discrepancies across frames. They considered a Recurrent Convolutional Network similar to~\cite{2018_David_AVSS}, trained end-to-end instead of using a pre-trained model. Their proposed detection approach was tested using FaceForensics++ database~\cite{rossler2019faceforensics++}, achieving AUC results of 96.9\% and 96.3\% for the DeepFake and FaceSwap methods, respectively. 

Therefore, very good fake detection results are already achieved on databases of the \nth{1} generation, being an almost solved problem. But, what is the performance achieved on current Identity Swap databases of the \nth{2} generation? 

The present study provides an exhaustive analysis of both \nth{1} and \nth{2} DeepFake generations using state-of-the-art fake detectors. Two different approaches are considered to detect fake videos: \textit{i)} the traditional one followed in the literature and based on selecting the entire face as input to the fake detection system~\cite{tolosana2020SurveyFakes}, and \textit{ii)} a novel approach based on the selection of specific facial regions as input to the fake detection system. The main contributions of this study are as follow:
\begin{itemize}
\item An in-depth comparison in terms of performance among Identity Swap databases of the \nth{1} and \nth{2} generation. In particular, two different state-of-the-art fake detectors are considered: \textit{i)} Xception, and \textit{ii)} Capsule Network.  
\item An analysis of the discriminative power of the different facial regions between the \nth{1} and \nth{2} generations, and also between fake detectors.  
\end{itemize}

The analysis carried out in this study will benefit the research community for many different reasons: \textit{i)} insights for the proposal of more robust fake detectors, e.g., through the fusion of different facial regions depending on the scenario: light conditions, pose variations, and distance from the camera; and \textit{ii)} the improvement of the next generation of DeepFakes, focusing on the artifacts existing in specific facial regions.

The remainder of the paper is organised as follows. Sec.~\ref{proposedApproach} describes our proposed evaluation framework. Sec.~\ref{databases} summarises all databases considered in the experimental framework of this study. Sec.~\ref{experimentalProtocol} and \ref{experimentalResults} describe the experimental protocol and results achieved, respectively. Finally, Sec.~\ref{conclusions} draws the final conclusions and points out future research lines.

\section{Proposed Evaluation Framework}\label{proposedApproach}
Fig.~\ref{fig:abstract_figure} graphically summarises our evaluation framework. It comprises two main modules: \textit{i)} facial region segmentation, described in Sec.~\ref{facial_region_segmentation}, and \textit{ii)} fake detection systems, described in Sec.~\ref{fake_detection_system}.

\subsection{Facial Region Segmentation}\label{facial_region_segmentation}
Two different approaches are studied: \textit{i)} segmenting the entire face as input to the fake detection system, and \textit{ii)} segmenting only specific facial regions. 

\begin{figure}[t]
\begin{center}
   \includegraphics[width=0.8\linewidth]{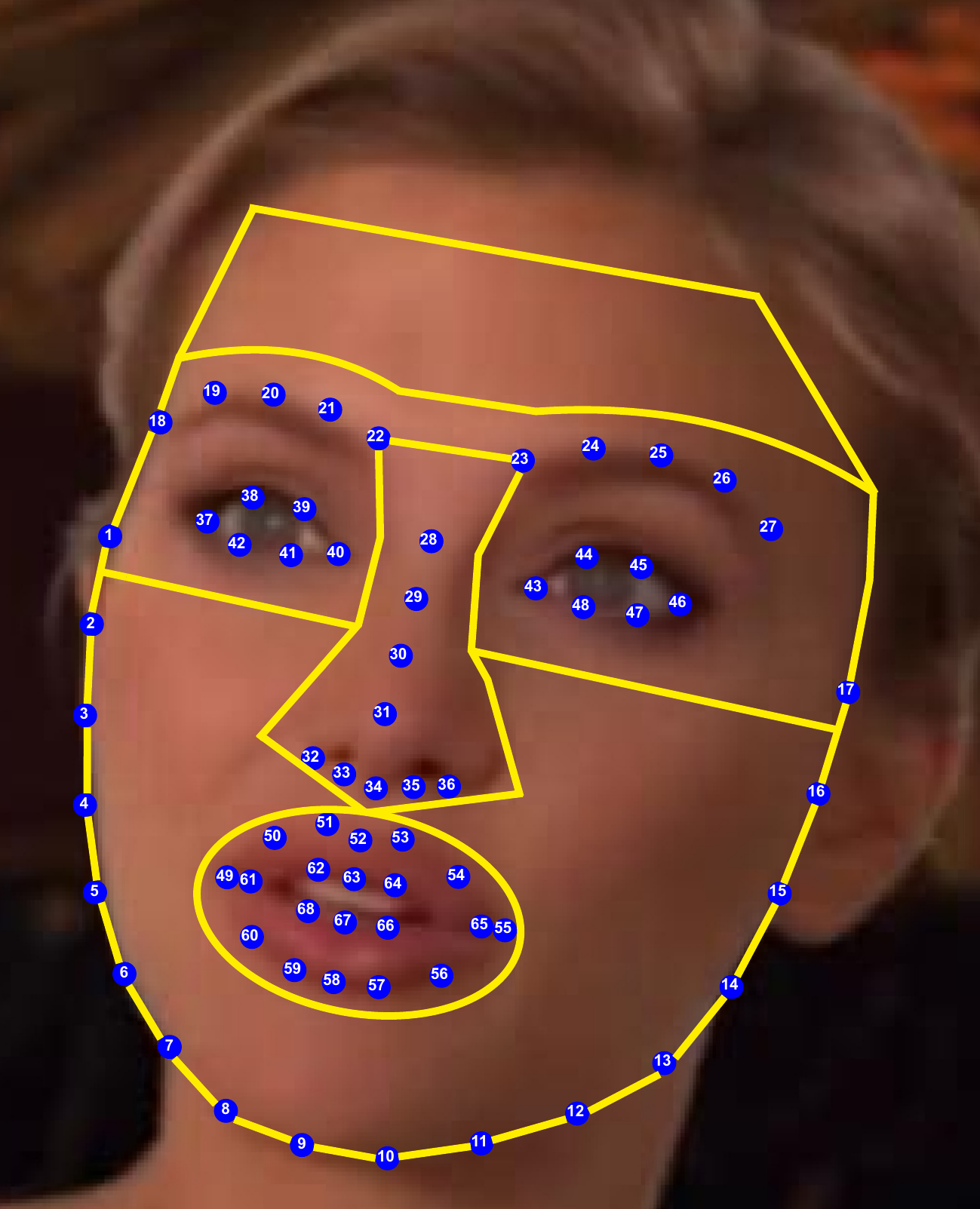}
\end{center}
   \caption{Example of the different facial regions (i.e., \textit{Eyes}, \textit{Nose}, \textit{Mouth}, and \textit{Rest}) extracted using the 68 facial landmarks provided by OpenFace2~\cite{baltrusaitis2018openface}.}
\label{fig:landmarks}
\end{figure}

Regarding the second approach, 4 different facial regions are selected: eyes, nose, mouth, and rest (i.e., the part of the face obtained after removing the eyes, nose, and mouth from the entire face). For the segmentation of each region, we consider the open-source toolbox OpenFace2~\cite{baltrusaitis2018openface}. This toolbox extracts 68 total landmarks for each face. Fig.~\ref{fig:landmarks} shows an example of the 68 landmarks (blue circles) extracted by OpenFace2 over a frame of the Celeb-DF database. It is important to highlight that OpenFace2 is robust against pose variations, distance from the camera, and light conditions, extracting reliable landmarks even for challenging databases such as the DFDC database~\cite{dolhansky2019deepfake}. The specific key landmarks considered to extract each facial region are as follow:

\begin{itemize}
\item \textit{Eyes}: using landmark points from 18 to 27 (top of the mask), and using landmarks 1, 2, 16, and 17 (bottom of the mask).
\item \textit{Nose}: using landmark points 22, 23 (top of the mask), from 28 to 36 (line and bottom of the nose), and 40, 43 (width of the middle-part of the nose).
\item \textit{Mouth}: using landmark points 49, 51-53, 55, and 57-59 to build a circular/elliptical mask. 
\item \textit{Rest}: extracted after removing eyes, nose, and mouth masks from the entire face.
\end{itemize}


Each facial region is highlighted by yellow lines in Fig.~\ref{fig:landmarks}. Once each facial region is segmented, the remaining part of the face is discarded (black background as depicted in Fig.~\ref{fig:abstract_figure}). Also, for each facial region, we keep the same image size and resolution as the original face image to perform a fair evaluation among facial regions and the entire face, avoiding therefore the influence of other pre-processing aspects such as interpolation.

\subsection{Fake Detection Systems}\label{fake_detection_system}
Two different state-of-the-art fake detection approaches are considered in our evaluation framework:

\begin{itemize}
\item \textit{Xception}~\cite{chollet2017xception}: this network has achieved very good fake detection results in recent studies~\cite{rossler2019faceforensics++,dolhansky2019deepfake,2019_Arxiv_GANRemoval_Tolosana,Jain2019facialManipulation}. Xception is a CNN architecture inspired by Inception~\cite{szegedy2015going}, where Inception modules have been replaced with depthwise separable convolutions. In our evaluation framework, we follow the same training approach considered in~\cite{rossler2019faceforensics++}: \textit{i)} we first consider the Xception model pre-trained with ImageNet~\cite{deng2009imagenet}, \textit{ii)} we change the last fully-connected layer of the ImageNet model by a new one (two classes, real or fake), \textit{iii)} we fix all weights up to the final fully-connected layer and re-train the network for few epochs, and finally \textit{iv)} we train the whole network for 20 more epochs and choose the best performing model based on validation accuracy.
\item \textit{Capsule Network}~\cite{nguyen2019use}: we consider the same detection approach proposed by Nguyen \textit{et al.}, which is publicly available in GitHub\footnote{\url{https://github.com/nii-yamagishilab/Capsule-Forensics-v2}}. It is based on the combination of traditional CNN and recent Capsule Networks, which require fewer parameters to train compared with traditional CNN~\cite{capsuleNetworks}. In particular, the authors proposed to use part of the VGG19 model pre-trained with ImageNet database for the feature extractor (from the first layer to the third max-pooling layer). The output of this pre-trained part is concatenated with 10 primary capsules and finally 2 output capsules (real and fake). In our evaluation framework, we train only the capsules following the procedure described in~\cite{nguyen2019use}.

\end{itemize}

Finally, as shown in Fig.~\ref{fig:abstract_figure}, it is important to highlight that we train a specific fake detector per database and facial region.

\begin{table}[t]
\centering
\caption{Identity swap publicly available databases of the \nth{1} and \nth{2} generations considered in our experimental framework.}
\label{table:databases_faceSwap}
\scalebox{1}{
\begin{tabular}{ccc}
\multicolumn{3}{c}{\textbf{\nth{1} Generation}} \\ \hline
\textit{Database}                                                              & \textit{Real Videos} & \textit{Fake Videos}                                                      \\ \hline
\begin{tabular}[c]{@{}c@{}}UADFV (2018)\\ \cite{li2018ictu}\end{tabular}              & 49 (Youtube)         & 49 (FakeApp)                                                              \\ 
\begin{tabular}[c]{@{}c@{}}FaceForensics++ (2019)\\ \cite{rossler2019faceforensics++}\end{tabular}   & 1,000 (Youtube)       & \begin{tabular}[c]{@{}c@{}}1,000 (FaceSwap)\end{tabular} \\
                                   &             &             \\   
\multicolumn{3}{c}{\textbf{\nth{2} Generation}} \\ \hline
\textit{Database}                                                              & \textit{Real Videos} & \textit{Fake Videos}                                                          \\ \hline
\begin{tabular}[c]{@{}c@{}}Celeb-DF (2019)\\ \cite{li2019celebdf}\end{tabular}          & 408 (Youtube)        & 795 (DeepFake)                                                            \\ 
\begin{tabular}[c]{@{}c@{}}DFDC Preview (2019)\\ \cite{dolhansky2019deepfake}\end{tabular}     & 1,131 (Actors)        & 4,119 (Unknown)                                                            \\ 
\end{tabular}
}
\end{table}

\section{Databases}\label{databases}
Four different public databases are considered in the experimental framework of this study. In particular, two databases of the \nth{1} generation (UADFV and FaceForensics++) and two recent databases of the \nth{2} generation (Celeb-DF and DFDC). Table~\ref{table:databases_faceSwap} summarises their main features.

\subsection{UADFV}\label{uadfv_database}
The UADFV database~\cite{li2018ictu} comprises 49 real videos downloaded from Youtube, which were used to create 49 fake videos through the FakeApp mobile application\footnote{\url{https://fakeapp.softonic.com/}}, swapping in all of them the original face with the face of the actor Nicolas Cage. Therefore, only one identity is considered in all fake videos. Each video represents one individual, with a typical resolution of 294$\times$500 pixels, and 11.14 seconds on average.

\subsection{FaceForensics++}\label{faceForensiccs_database}
The FaceForensics++ database~\cite{rossler2019faceforensics++} was introduced in 2019 as an extension of the original FaceForensics~\cite{rossler2018faceforensics}, which was focused only on Expression Swap manipulations. FaceForensics++ contains 1,000 real videos extracted from Youtube. Fake videos were generated using both computer graphics and deep learning approaches (1,000 fake videos for each approach). In this study we focus on the computer graphics approach where fake videos were created using the publicly available FaceSwap algorithm\footnote{\url{https://github.com/MarekKowalski/FaceSwap}}. This algorithm consists of face alignment, Gauss Newton optimization and image blending to swap the face of the source person to the target person.

\subsection{Celeb-DF}\label{uadfv_database}
The aim of the Celeb-DF database~\cite{li2019celebdf} was to generate fake videos of better visual quality compared with their original UADFV database. This database consists of 408 real videos extracted from Youtube, corresponding to interviews of 59 celebrities with a diverse distribution in terms of gender, age, and ethnic group. In addition, these videos exhibit a large range of variations in aspects such as the face sizes (in pixels), orientations, lighting conditions, and backgrounds. Regarding fake videos, a total of 795 videos were created using DeepFake technology, swapping faces for each pair of the 59 subjects. The final videos are in MPEG4.0 format.

\subsection{DFDC}\label{DFDC_database}
The DFDC database~\cite{dolhansky2019deepfake} is one of the latest public databases, released by Facebook in collaboration with other companies and academic institutions such as Microsoft, Amazon, and the MIT. In the present study we consider the DFDC preview dataset consisting of 1,131 real videos from 66 paid actors, ensuring realistic variability in gender, skin tone, and age. It is important to remark that no publicly available data or data from social media sites were used to create this dataset, unlike other popular databases. Regarding fake videos, a total of 4,119 videos were created using two different unknown approaches for fakes generation. Fake videos were generated by swapping subjects with similar appearances, i.e., similar facial attributes such as skin tone, facial hair, glasses, etc. After a given pairwise model was trained on two identities, they swapped each identity onto the other’s videos.  

It is important to highlight that the DFDC database considers different acquisition scenarios (i.e., indoors and outdoors), light conditions (i.e., day, night, etc.), distances from the person to the camera, and pose variations, among others.

\begin{table*}[t]
\centering
\caption{Fake detection performance results in terms of EER (\%) and AUC (\%) over the final evaluation datasets. Two approaches are considered as input to the fake detection systems: \textit{i)} selecting the entire face (\textit{Face}), and \textit{ii)} selecting specific facial regions (\textit{Eyes}, \textit{Nose}, \textit{Mouth}, \textit{Rest}). \nth{1} generation databases: UADFV and FaceForensic++. \nth{2} generation databases: Celeb-DF and DFDC. For each database, we remark in \textbf{bold} the best fake detection results, and in {\color[HTML]{00BFFF}blue} and {\color[HTML]{ED872D}orange} the facial regions that provide the {\color[HTML]{00BFFF}best} and {\color[HTML]{ED872D}worst} results, respectively.\vspace{3mm}}
\label{table:performance_results}
\scalebox{0.98}{
\begin{tabular}{ccccccccccc}
\cline{2-11}
\multicolumn{1}{c|}{\textbf{\textit{Xception}}}                                                                         & \multicolumn{2}{c|}{\textit{Face}}       & \multicolumn{2}{c|}{\textit{Eyes}}       & \multicolumn{2}{c|}{\textit{Nose}}       & \multicolumn{2}{c|}{\textit{Mouth}}      & \multicolumn{2}{c|}{\textit{Rest}}       \\ \cline{2-11} 
\multicolumn{1}{c|}{}                                                                          & EER (\%) & \multicolumn{1}{c|}{AUC (\%)} & EER (\%) & \multicolumn{1}{c|}{AUC (\%)} & EER (\%) & \multicolumn{1}{c|}{AUC (\%)} & EER (\%) & \multicolumn{1}{c|}{AUC (\%)} & EER (\%) & \multicolumn{1}{c|}{AUC (\%)} \\ \hline
\multicolumn{1}{c|}{\begin{tabular}[c]{@{}c@{}}UADFV (2018)\\ \cite{li2018ictu}\end{tabular}}           & \textbf{1.00}     & \multicolumn{1}{c|}{\textbf{100}}   & \color[HTML]{00BFFF} 2.20     & \multicolumn{1}{c|}{ \color[HTML]{00BFFF}99.70}    & \color[HTML]{ED872D}13.50    & \multicolumn{1}{c|}{\color[HTML]{ED872D}94.70}    & 12.50    & \multicolumn{1}{c|}{95.40}    & 7.90     & \multicolumn{1}{c|}{97.30}    \\
\multicolumn{1}{c|}{\begin{tabular}[c]{@{}c@{}}FaceForensics++ (2019)\\ \cite{rossler2019faceforensics++}\end{tabular}} & \textbf{3.31}     & \multicolumn{1}{c|}{\textbf{99.40}}    & 14.23    & \multicolumn{1}{c|}{92.70}    & 21.97    & \multicolumn{1}{c|}{86.30}    &  \color[HTML]{00BFFF}13.77    & \multicolumn{1}{c|}{ \color[HTML]{00BFFF}93.90}    & \color[HTML]{ED872D}22.37    & \multicolumn{1}{c|}{\color[HTML]{ED872D}85.50}    \\ \hline
                                                                                              &          &                               &          &                               &          &                               &          &                               &          &                               \\ \hline
\multicolumn{1}{c|}{\begin{tabular}[c]{@{}c@{}}Celeb-DF (2019)\\ \cite{li2019celebdf}\end{tabular}}        &     \textbf{28.55}     & \multicolumn{1}{c|}{\textbf{83.60}}    &          \color[HTML]{00BFFF}29.40 & \multicolumn{1}{c|}{ \color[HTML]{00BFFF}77.30}    & 38.46    & \multicolumn{1}{c|}{64.90}    & 39.37    & \multicolumn{1}{c|}{65.10}    & \color[HTML]{ED872D}43.55    & \multicolumn{1}{c|}{ \color[HTML]{ED872D}60.10}    \\
\multicolumn{1}{c|}{\begin{tabular}[c]{@{}c@{}}DFDC Preview (2019)\\ \cite{dolhansky2019deepfake}\end{tabular}}    & \textbf{17.55}    & \multicolumn{1}{c|}{\textbf{91.17}}    &  \color[HTML]{00BFFF}23.82    & \multicolumn{1}{c|}{ \color[HTML]{00BFFF}83.90}    & 26.80    & \multicolumn{1}{c|}{81.50}    & 27.59    & \multicolumn{1}{c|}{79.50}    & \color[HTML]{ED872D}29.94    & \multicolumn{1}{c|}{\color[HTML]{ED872D}76.50}    \\ \hline
\end{tabular}
}
\end{table*}

\begin{table*}[t]
\centering
\vspace{3mm}
\scalebox{0.98}{
\begin{tabular}{ccccccccccc}
\cline{2-11}
\multicolumn{1}{c|}{\textbf{\textit{Capsule Network}}}                                                                         & \multicolumn{2}{c|}{\textit{Face}}       & \multicolumn{2}{c|}{\textit{Eyes}}       & \multicolumn{2}{c|}{\textit{Nose}}       & \multicolumn{2}{c|}{\textit{Mouth}}      & \multicolumn{2}{c|}{\textit{Rest}}       \\ \cline{2-11} 
\multicolumn{1}{c|}{}                                                                          & EER (\%) & \multicolumn{1}{c|}{AUC (\%)} & EER (\%) & \multicolumn{1}{c|}{AUC (\%)} & EER (\%) & \multicolumn{1}{c|}{AUC (\%)} & EER (\%) & \multicolumn{1}{c|}{AUC (\%)} & EER (\%) & \multicolumn{1}{c|}{AUC (\%)} \\ \hline
\multicolumn{1}{c|}{\begin{tabular}[c]{@{}c@{}}UADFV (2018)\\ \cite{li2018ictu}\end{tabular}}           & 2.00     & \multicolumn{1}{c|}{99.90}   & \color[HTML]{00BFFF} \textbf{0.28}     & \multicolumn{1}{c|}{ \color[HTML]{00BFFF}\textbf{100}}    & 3.92    & \multicolumn{1}{c|}{99.30}    & 3.20    & \multicolumn{1}{c|}{99.56}    & \color[HTML]{ED872D}12.30     & \multicolumn{1}{c|}{\color[HTML]{ED872D}94.83}    \\
\multicolumn{1}{c|}{\begin{tabular}[c]{@{}c@{}}FaceForensics++ (2019)\\ \cite{rossler2019faceforensics++}\end{tabular}} & \textbf{2.75}     & \multicolumn{1}{c|}{\textbf{99.52}}    & 10.29    & \multicolumn{1}{c|}{95.32}    & 17.51    & \multicolumn{1}{c|}{90.09}    &  \color[HTML]{00BFFF}9.66    & \multicolumn{1}{c|}{ \color[HTML]{00BFFF}96.18}    & \color[HTML]{ED872D}21.58    & \multicolumn{1}{c|}{\color[HTML]{ED872D}86.61}    \\ \hline
                                                                                              &          &                               &          &                               &          &                               &          &                               &          &                               \\ \hline
\multicolumn{1}{c|}{\begin{tabular}[c]{@{}c@{}}Celeb-DF (2019)\\ \cite{li2019celebdf}\end{tabular}}        &     \textbf{24.29}     & \multicolumn{1}{c|}{\textbf{82.46}}    &          \color[HTML]{00BFFF}30.58 & \multicolumn{1}{c|}{ \color[HTML]{00BFFF}76.64}    & 37.39    & \multicolumn{1}{c|}{66.24}    & 35.36    & \multicolumn{1}{c|}{67.75}    & \color[HTML]{ED872D}36.64    & \multicolumn{1}{c|}{ \color[HTML]{ED872D}68.56}    \\
\multicolumn{1}{c|}{\begin{tabular}[c]{@{}c@{}}DFDC Preview (2019)\\ \cite{dolhansky2019deepfake}\end{tabular}}    & \textbf{21.39}    & \multicolumn{1}{c|}{\textbf{87.45}}    &  \color[HTML]{00BFFF}25.06    & \multicolumn{1}{c|}{ \color[HTML]{00BFFF}83.12}    & 26.53    & \multicolumn{1}{c|}{81.50}    & 27.92    & \multicolumn{1}{c|}{78.14}    & \color[HTML]{ED872D}32.56    & \multicolumn{1}{c|}{\color[HTML]{ED872D}72.42}    \\ \hline
\end{tabular}
}
\end{table*}

\section{Experimental Protocol}\label{experimentalProtocol}
All databases have been divided into non-overlapping datasets, development ($\simeq$ 80\% of the identities) and evaluation ($\simeq$ 20\% of the identities). It is important to remark that each dataset comprises videos from different identities (both real and fake), unlike some previous studies. This aspect is very important in order to perform a fair evaluation and predict the generalisation ability of the fake detection systems against unseen identities. For example, for the UADFV database, all real and fake videos related to the identity of Donald Trump were considered only for the final evaluation of the models. For the FaceForensics++ database, we consider 860 development videos and 140 evaluation videos per class (real/fake) as proposed in~\cite{rossler2019faceforensics++}, selecting different identities in each dataset (one fake video is provided for each identity). For the DFDC Preview database, we follow the same experimental protocol proposed in~\cite{dolhansky2019deepfake} as the authors already considered this concern. Finally, for the Celeb-DF database, we consider real/fake videos of 40 and 19 different identities for the development and evaluation datasets, respectively.

\section{Experimental Results}\label{experimentalResults}
Two experiments are considered: \textit{i)} Sec.~\ref{entire_face_analysis} considers the traditional scenario of feeding the fake detectors with the entire face, and \textit{ii)} Sec.~\ref{facial_region_analysis} analyses the discriminative power of each facial region. Finally, we compare in Sec.~\ref{state_art} the results achieved in this study with the state of the art.

\subsection{Entire Face Analysis}\label{entire_face_analysis}
Table~\ref{table:performance_results} shows the fake detection performance results achieved in terms of Equal Error Rate (EER) and AUC over the final evaluation datasets of both \nth{1} and \nth{2} generations of fake videos. The results achieved using the entire face are indicated as \textit{Face}. For each database and fake detection approach, we remark in \textbf{bold} the best performance results achieved.

Analysing the fake videos of the \nth{1} generation, AUC values close to 100\% are achieved, proving how easy it is for both systems to detect fake videos of the \nth{1} generation. In terms of EER, higher fake detection errors are achieved when using the FaceForensics++ database (around 3\% EER), proving to be more challenging than the UADFV database.

\begin{table}[t]
\centering
\caption{Fake detection results in terms of EER (\%) using Xception over the final evaluation dataset of Celeb-DF. Two scenarios are analysed regarding whether the same identities are used for the development and final evaluation of the detectors or not. In both scenarios, different videos (real and fake) are considered in each dataset.}
\resizebox{0.48\textwidth}{!}{%
\begin{tabular}{c|ccccc}
                              & \textit{Face} & \textit{Eyes} & \textit{Nose} & \textit{Mouth} & \textit{Rest} \\ \hline
\textit{Same} identities      & \textbf{5.66}         & 12.06         & 23.44         & 17.81          & 21.58         \\
\textit{Different} identities & \textbf{28.55}         & 29.40         & 38.46         & 39.37          & 43.55         \\ \hline
\end{tabular}\label{table:identities}
}
\end{table}

Regarding the DeepFake databases of the \nth{2} generation, a high performance degradation is observed in both fake detectors when using Celeb-DF and DFDC databases. In particular, an average 23.05\% EER is achieved for Xception whereas for Capsule Network, the average EER is 22.84\%. As a result, an average absolute worsening of around 20\% EER is produced for both fake detectors compared with the databases of the \nth{1} generation. This degradation is specially substantial for the Celeb-DF database, with EER values of 28.55\% and 24.29\% for Xception and Capsule Network fake detectors, respectively. These results prove the higher realism achieved in the \nth{2} in comparison with the \nth{1} DeepFake generation.

Finally, we would like to highlight the importance of selecting different identities (not only videos) for the development and final evaluation of the fake detectors, as we have done in our experimental framework. As an example of how relevant this aspect is, Table~\ref{table:identities} shows the detection performance results achieved using Xception for the \textit{Same} and \textit{Different} identities between development and evaluation of Celeb-DF. As can be seen, much better results are obtained for the scenario of considering the \textit{Same} identities, up to 5 times better compared with the \textit{Different} identities scenario. The \textit{Same} identities scenario generates a misleading result because the network is learning intrinsic features from the identities, not the key features to distinguish among real and fake videos. Therefore, poor results are expected to be achieved when testing with other identities. This is a key aspect not considered in the experimental protocol of many previous studies.

\subsection{Facial Regions Analysis}\label{facial_region_analysis}
\begin{figure*}[!]
\begin{center}
   \includegraphics[width=\linewidth]{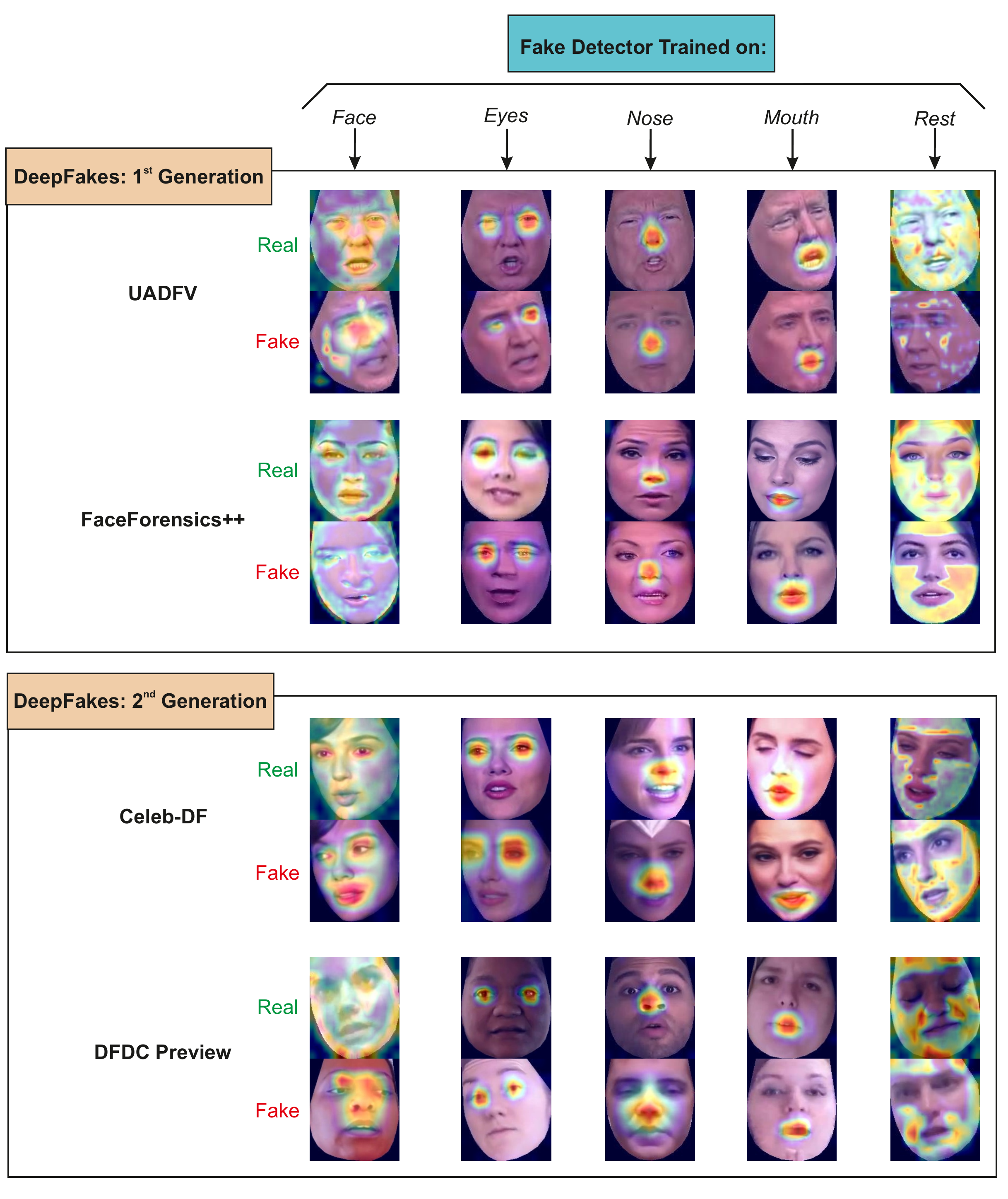}
\end{center}
   \caption{Real and fake image examples of the DeepFake video databases evaluated in the present paper with their corresponding Grad-CAM heatmaps, representing the facial features most useful for each fake detector (i.e., \textit{Face}, \textit{Eyes}, \textit{Nose}, \textit{Mouth}, and \textit{Rest}).}\label{fig:heatmaps}
\end{figure*}

Table~\ref{table:performance_results} also includes the results achieved for each specific facial region: \textit{Eyes}, \textit{Nose}, \textit{Mouth}, and \textit{Rest}. For each database and fake detection approach, we remark in {\color[HTML]{00BFFF}blue} and {\color[HTML]{ED872D}orange} the facial regions that provide the {\color[HTML]{00BFFF}best} and {\color[HTML]{ED872D}worst} results, respectively. It is important to remark that a separate fake detection model is trained for each facial region and database. In addition, we also visualise in Fig.~\ref{fig:heatmaps} which part of the image is more important for the final decision, for both real and fake examples. We consider the popular heatmap visualisation technique Grad-CAM~\cite{selvaraju2017grad}. Similar Grad-CAM results are obtained for both Xception and Capsule Network.

In general, as shown in Table~\ref{table:performance_results}, the facial region \textit{Eyes} provides the best results whereas the \textit{Rest} (i.e., the remaining part of the face after removing eyes, nose, and mouth) provides the worst results. 

For the UADFV database, the \textit{Eyes} provides EER values close to the results achieved using the entire \textit{Face}. It is important to highlight the results achieved by the Capsule Network as in this case the fake detector based only on the \textit{Eyes} has outperformed the case of feeding the detector with the entire \textit{Face} (2.00\% vs. 0.28\% EER). The discriminative power of the \textit{Eyes} facial region was preliminary studied by Matern \textit{et al.} in~\cite{matern2019exploiting}, proposing features based on the missing reflection details of the eyes. Also, in this particular database, Xception achieves good results using the \textit{Rest} of the face, 7.90\% EER. This is produced due to the different colour contrast among the synthesised fake mask and real skin, and also to the visible boundaries of the fake mask. These aspects can be noticed in the examples included in Fig.~\ref{fig:heatmaps}. 

Regarding the FaceForensics++ database, the \textit{Mouth} is the facial region that achieves the best result for both Xception and Capsule Network with EER values of 13.77\% and 9.66\%. This is produced due to the lack of details in the teeth (blurred) and also the lip inconsistencies among the original face and the synthesised. Similar results are obtained when using the \textit{Eyes}. It is interesting to see in Fig.~\ref{fig:heatmaps} how the decision of the fake detection systems is mostly based on a single eye (the same happens in other databases such as UADFV). Finally, the fake detection system based on the \textit{Rest} of the face provides the worst result, EER values of 22.37\% and 21.58\% for Xception and Capsule Network, respectively. This may happen because both colour contrast and visible boundaries were further improved in FaceForensics++ compared with the UADFV database.  

It is also interesting to analyse the ability of each approach for the detection of fake videos of the \nth{1} generation. In general, much better results are obtained using Capsule Networks compared with Xception. For example, regarding the UADFV database, EER absolute improvements of 1.92\%, 9.58\%, and 9.30\% are obtained for the \textit{Eyes}, \textit{Nose}, and \textit{Mouth}, respectively.

\begin{table*}[t]
\centering
\caption{Comparison in terms of AUC (\%) of different state-of-the-art fake detectors with the present study. The best results achieved for each database are remarked in \textbf{bold}. Results in \textit{italics} indicate that the evaluated database was not used for training~\cite{li2019celebdf}.}
\label{table:comparison_state_art}
\resizebox{\textwidth}{!}{%
\begin{tabular}{ccccccc}
\multirow{2}{*}{Study}         & \multirow{2}{*}{Method}                 & \multirow{2}{*}{Classifiers} & \multicolumn{4}{c}{AUC Results (\%)}                                       \\ \cline{4-7} 
                               &                                         &                              & UADFV~\cite{li2018ictu} & FF++~\cite{rossler2019faceforensics++}  & Celeb-DF~\cite{li2019celebdf} & DFDC~\cite{dolhansky2019deepfake} \\ \hline \hline
Yang \textit{et al.}~\cite{yang2019exposing}                    & Head Pose Features                      & SVM                          & 89.0             & \textit{47.3}   & \textit{54.6}       & \textit{55.9}   \\
Li \textit{et al.}~\cite{li2019celebdf}                      & Face Warping Features                   & CNN                          & 97.7             & \textit{93.0}   & \textit{64.6}       & \textit{75.5}   \\
Afchar \textit{et al.}~\cite{afchar2018mesonet}                  & Mesoscopic Features                     & CNN                          & \textit{84.3}    & 84.7   & \textit{54.8}       & \textit{75.3}   \\
Sabir \textit{et al.}~\cite{2019_Sabir}                   & Image + Temporal Features               & CNN + RNN                    & -                & 96.3            & -                   & -               \\
Dang \textit{et al.}~\cite{Jain2019facialManipulation}                    & Deep Learning Features                  & CNN + Attention Mechanism    & 98.4             & -               & \textit{71.2}       & -               \\ \hline 
\multirow{2}{*}{Present Study} & \multirow{2}{*}{Deep Learning Features} & Xception~\cite{chollet2017xception}                     & \textbf{100}     & 99.4           & \textbf{83.6}      & \textbf{91.1}  \\
                               &                                         & Capsule Network~\cite{nguyen2019use}    & \textbf{100}            & \textbf{99.5}  & 82.4               & 87.4           \\ \hline
\end{tabular}
}
\end{table*}

Analysing the Celeb-DF database of the \nth{2} generation, the best results for local regions are achieved when using the \textit{Eyes} of the face, with EER values around 30\%, similar to using the entire \textit{Face} for Xception. It is important to remark that this EER is over 13 times higher than the original 2.20\% and 0.28\% EERs achieved by Xception and Capsule Network on the UADFV. Similar poor detection results, around 40\% EER, are obtained when using other facial regions, being one of the most challenging databases nowadays. Fig.~\ref{fig:heatmaps} depicts some fake examples of Celeb-DF, showing very realistic features such as the colour contrast, boundaries of the mask, quality of the eyes, teeth, nose, etc. 

Regarding the DFDC database, better detection results are obtained compared with the Celeb-DF database. In particular, the facial region \textit{Eyes} also provides the best results with EER values of 23.82\% and 25.06\%, an absolute improvement of 5.58\% and 5.52\% EER compared with the \textit{Eyes} facial region of Celeb-DF. Despite this performance improvement, the EER is still much worse compared with the databases of the \nth{1} generation.

To summarise this section, we have observed significant improvements in the realism of DeepFakes of the \nth{2} in comparison with the \nth{1} generation for some specific facial regions. In particular, for the \textit{Nose}, \textit{Mouth}, and the edge of the face (\textit{Rest}). This realism provokes a lot of fake detection errors even for the advanced detectors explored in the present paper, which result in EER values between 24\% and 44\% depending on the database. The quality of the \textit{Eyes} has also been improved, but it is still the facial region most useful to detect fake images, as depicted in Fig.~\ref{fig:heatmaps}.

\subsection{Comparison with the State of the Art}\label{state_art}
Finally, we compare in Table~\ref{table:comparison_state_art} the AUC results achieved in the present study with the state of the art. Different methods are considered to detect fake videos: head pose variations~\cite{yang2019exposing}, face warping artifacts~\cite{li2019celebdf}, mesoscopic features~\cite{afchar2018mesonet}, image and temporal features~\cite{2019_Sabir}, and pure deep learning features in combination with attention mechanisms~\cite{Jain2019facialManipulation}. The best results achieved for each database are remarked in \textbf{bold}. Results in \textit{italics} indicate that the evaluated database was not used for training. These results are extracted from~\cite{li2019celebdf}.

Note that the comparison in Table~\ref{table:comparison_state_art} is not always under the same datasets and protocols, therefore it must be interpreted with care. Despite of that, it is patent that both Xception and Capsule Network fake detectors have achieved state-of-the-art results in all databases. In particular, for Celeb-DF and DFDC, Xception obtains the best results whereas for FaceForensics++, Capsule Network is the best. The same good results are obtained by both detectors on UADFV.

\section{Conclusions}\label{conclusions}
In this study we have performed an exhaustive analysis of the DeepFakes evolution, focusing on facial regions and fake detection performance. Popular databases such as UADFV and FaceForensics++ of the \nth{1} generation, as well as the latest databases such as Celeb-DF and DFDC of the \nth{2} generation, are considered in the analysis.

Two different approaches have been followed in our evaluation framework to detect fake videos: \textit{i)} selecting the entire face as input to the fake detection system, and \textit{ii)} selecting specific facial regions such as the eyes or nose, among others, as input to the fake detection system.

Regarding the fake detection performance, we highlight the very poor results achieved in the latest DeepFake video databases of the \nth{2} generation with EER values around 20-30\%, compared with the EER values of the \nth{1} generation ranging from 1\% to 3\%. In addition, we remark the significant improvements in the realism achieved at image level in some facial regions such as the nose, mouth, and edge of the face in DeepFakes of the \nth{2} generation, resulting in fake detection results between 24\% and 44\% EERs.

The analysis carried out in this study provides useful insights for the research community, e.g.: \textit{i)} for the proposal of more robust fake detectors, e.g., through the fusion of different facial regions depending on the scenario: light conditions, pose variations, and distance from the camera; and \textit{ii)} the improvement of the next generation of DeepFakes, focusing on the artifacts existing in specific facial regions.

\section*{Acknowledgments}
This work has been supported by projects: PRIMA (H2020-MSCA-ITN-2019-860315), TRESPASS-ETN (H2020-MSCA-ITN-2019-860813), BIBECA (MINECO/FEDER RTI2018-101248-B-I00), and Accenture. Ruben Tolosana is supported by Consejer\'ia de Educaci\'on, Juventud y Deporte de la Comunidad de Madrid y Fondo Social Europeo.

%

{
\bibliographystyle{IEEEtran}
\bibliography{mybibfile}

\begin{thebibliography}{10}
\providecommand{\url}[1]{#1}
\csname url@samestyle\endcsname
\providecommand{\newblock}{\relax}
\providecommand{\bibinfo}[2]{#2}
\providecommand{\BIBentrySTDinterwordspacing}{\spaceskip=0pt\relax}
\providecommand{\BIBentryALTinterwordstretchfactor}{4}
\providecommand{\BIBentryALTinterwordspacing}{\spaceskip=\fontdimen2\font plus
\BIBentryALTinterwordstretchfactor\fontdimen3\font minus
  \fontdimen4\font\relax}
\providecommand{\BIBforeignlanguage}[2]{{%
\expandafter\ifx\csname l@#1\endcsname\relax
\typeout{** WARNING: IEEEtran.bst: No hyphenation pattern has been}%
\typeout{** loaded for the language `#1'. Using the pattern for}%
\typeout{** the default language instead.}%
\else
\language=\csname l@#1\endcsname
\fi
#2}}
\providecommand{\BIBdecl}{\relax}
\BIBdecl

\bibitem{tolosana2020SurveyFakes}
R.~Tolosana, R.~Vera-Rodriguez, J.~Fierrez, A.~Morales, and J.~Ortega-Garcia,
  ``{DeepFakes and Beyond: A Survey of Face Manipulation and Fake Detection},''
  \emph{Information Fusion}, 2020.

\bibitem{verdoliva2020media}
L.~Verdoliva, ``{Media Forensics and DeepFakes: an Overview},'' \emph{arXiv
  preprint arXiv:2001.06564}, 2020.

\bibitem{TED_news_concerns}
\BIBentryALTinterwordspacing
D.~Citron, ``{How DeepFake Undermine Truth and Threaten Democracy},'' 2019.
  [Online]. Available: \url{https://www.ted.com}
\BIBentrySTDinterwordspacing

\bibitem{BBC_doubleVideos}
\BIBentryALTinterwordspacing
R.~Cellan-Jones, ``{Deepfake Videos Double in Nine Months},'' 2019. [Online].
  Available: \url{https://www.bbc.com/news/technology-49961089}
\BIBentrySTDinterwordspacing

\bibitem{li2018ictu}
Y.~Li, M.~Chang, and S.~Lyu, ``{In Ictu Oculi: Exposing AI Generated Fake Face
  Videos by Detecting Eye Blinking},'' in \emph{Proc. IEEE International
  Workshop on Information Forensics and Security}, 2018.

\bibitem{li2019celebdf}
Y.~Li, X.~Yang, P.~Sun, H.~Qi, and S.~Lyu, ``{Celeb-DF: A Large-Scale
  Challenging Dataset for DeepFake Forensics},'' in \emph{Proc. IEEE/CVF
  Conference on Computer Vision and Pattern Recognition}, 2020.

\bibitem{dolhansky2019deepfake}
B.~Dolhansky, R.~Howes, B.~Pflaum, N.~Baram, and C.~C. Ferrer, ``{The Deepfake
  Detection Challenge (DFDC) Preview Dataset},'' \emph{arXiv preprint
  arXiv:1910.08854}, 2019.

\bibitem{yang2019exposing}
X.~Yang, Y.~Li, and S.~Lyu, ``{Exposing Deep Fakes Using Inconsistent Head
  Poses},'' in \emph{Proc. International Conference on Acoustics, Speech and
  Signal Processing}, 2019.

\bibitem{king2009dlib}
D.~King, ``{DLib-ML: A Machine Learning Toolkit},'' \emph{Journal of Machine
  Learning Research}, vol.~10, pp. 1755--1758, 2009.

\bibitem{Li2019CVPR}
Y.~Li and S.~Lyu, ``{Exposing DeepFake Videos By Detecting Face Warping
  Artifacts},'' in \emph{Proc. IEEE/CVF Conference on Computer Vision and
  Pattern Recognition Workshops}, 2019.

\bibitem{korshunov2018deepfakes}
P.~Korshunov and S.~Marcel, ``{Deepfakes: a New Threat to Face Recognition?
  Assessment and Detection},'' \emph{arXiv preprint arXiv:1812.08685}, 2018.

\bibitem{2019_Agarwal}
S.~Agarwal and H.~Farid, ``{Protecting World Leaders Against Deep Fakes},'' in
  \emph{Proc. IEEE/CVF Conference on Computer Vision and Pattern Recognition
  Workshops}, 2019.

\bibitem{2019_Sabir}
E.~Sabir, J.~Cheng, A.~Jaiswal, W.~AbdAlmageed, I.~Masi, and P.~Natarajan,
  ``{Recurrent Convolutional Strategies for Face Manipulation Detection in
  Videos},'' in \emph{Proc. IEEE/CVF Conference on Computer Vision and Pattern
  Recognition Workshops}, 2019.

\bibitem{2018_David_AVSS}
D.~G\"uera and E.~Delp, ``{Deepfake Video Detection Using Recurrent Neural
  Networks},'' in \emph{Proc. International Conference on Advanced Video and
  Signal Based Surveillance}, 2018.

\bibitem{rossler2019faceforensics++}
A.~R{\"o}ssler, D.~Cozzolino, L.~Verdoliva, C.~Riess, J.~Thies, and
  M.~Nie{\ss}ner, ``{FaceForensics++: Learning to Detect Manipulated Facial
  Images},'' in \emph{Proc. IEEE/CVF International Conference on Computer
  Vision}, 2019.

\bibitem{baltrusaitis2018openface}
T.~Baltrusaitis, A.~Zadeh, Y.~Lim, and L.~Morency, ``{OpenFace 2.0: Facial
  Behavior Analysis Toolkit},'' in \emph{Proc. International Conference on
  Automatic Face \& Gesture Recognition}, 2018.

\bibitem{chollet2017xception}
F.~Chollet, ``{Xception: Deep Learning with Depthwise Separable
  Convolutions},'' in \emph{Proc. IEEE/CVF Conference on Computer Vision and
  Pattern Recognition}, 2017.

\bibitem{2019_Arxiv_GANRemoval_Tolosana}
J.~Neves, R.~Tolosana, R.~Vera-Rodriguez, V.~Lopes, H.~Proen\c{c}a, and
  J.~Fierrez, ``{GANprintR: Improved Fakes and Evaluation of the State of the
  Art in Face Manipulation Detection},'' \emph{IEEE Journal of Selected Topics
  in Signal Processing}, 2020.

\bibitem{Jain2019facialManipulation}
H.~Dang, F.~Liu, J.~Stehouwer, X.~Liu, and A.~Jain, ``{On the Detection of
  Digital Face Manipulation},'' in \emph{Proc. IEEE/CVF Conference on Computer
  Vision and Pattern Recognition}, 2020.

\bibitem{szegedy2015going}
C.~Szegedy, W.~Liu, Y.~Jia, P.~Sermanet, S.~Reed, D.~Anguelov, D.~Erhan,
  V.~Vanhoucke, and A.~Rabinovich, ``{Going Deeper with Convolutions},'' in
  \emph{Proc. IEEE/CVF Conference on Computer Vision and Pattern Recognition},
  2015.

\bibitem{deng2009imagenet}
J.~Deng, W.~Dong, R.~Socher, L.~Li, K.~Li, and L.~Fei-Fei, ``{ImageNet: A
  Large-Scale Hierarchical Image Database},'' in \emph{Proc. IEEE/CVF
  Conference on Computer Vision and Pattern Recognition}, 2009.

\bibitem{nguyen2019use}
{H.H. Nguyen, J. Yamagishi and I. Echizen}, ``{Use of a Capsule Network to
  Detect Fake Images and Videos},'' \emph{arXiv preprint arXiv:1910.12467},
  2019.

\bibitem{capsuleNetworks}
{G.E. Hinton, S. Sabour and N. Frosst}, ``{Matrix Capsules with EM Routing},''
  in \emph{Proc. International Conference on Learning Representations
  Workshop}, 2018.

\bibitem{rossler2018faceforensics}
A.~R{\"o}ssler, D.~Cozzolino, L.~Verdoliva, C.~Riess, J.~Thies, and
  M.~Nie{\ss}ner, ``{FaceForensics: A Large-Scale Video Dataset for Forgery
  Detection in Human Faces},'' \emph{arXiv preprint arXiv:1803.09179}, 2018.

\bibitem{selvaraju2017grad}
R.~Selvaraju, M.~Cogswell, A.~Das, R.~Vedantam, D.~Parikh, and D.~Batra,
  ``{Grad-CAM: Visual Explanations from Deep Networks via Gradient-based
  Localization},'' in \emph{Proc. IEEE International Conference on Computer
  Vision}, 2017.

\bibitem{matern2019exploiting}
F.~Matern, C.~Riess, and M.~Stamminger, ``{Exploiting Visual Artifacts to
  Expose DeepFakes and Face Manipulations},'' in \emph{Proc. IEEE Winter
  Applications of Computer Vision Workshops}, 2019.

\bibitem{afchar2018mesonet}
D.~Afchar, V.~Nozick, J.~Yamagishi, and I.~Echizen, ``{MesoNet: a Compact
  Facial Video Forgery Detection Network},'' in \emph{Proc. IEEE International
  Workshop on Information Forensics and Security}, 2018.

\end{thebibliography}
}

\end{document}